\documentclass[journal]{IEEEtran}
\usepackage{amsmath,amsfonts}
\usepackage{algorithmic}
\usepackage{algorithm}
\usepackage{array}
\usepackage[caption=false,font=normalsize,labelfont=sf,textfont=sf]{subfig}
\usepackage{textcomp}
\usepackage{stfloats}
\usepackage{url}
\usepackage{verbatim}
\usepackage{graphicx}
\usepackage{cite}

\usepackage{graphicx}
\usepackage{hyperref}
\usepackage{times}
\usepackage{algorithm}
\usepackage{algorithmic}
\usepackage{algorithm}
\usepackage{algorithmic}
\usepackage{amsfonts,amssymb}
\usepackage{amsmath,bm}
\usepackage{multirow} 
\usepackage{multicol} 
\usepackage{xcolor}
\usepackage{colortbl}
\definecolor{Gray}{gray}{0.93}
\definecolor{lightgray}{gray}{0.9} 
\definecolor{alertcolor}{RGB}{255, 224, 224} 
\usepackage{mathrsfs}
\usepackage{tabularx}
\usepackage{makecell}
\usepackage{amsfonts}
\usepackage{bbm}
\usepackage{arydshln}
\usepackage{booktabs}
\usepackage{bbding}
\usepackage{pifont}
\usepackage{soul, color, xcolor}
\soulregister\cite7  
\soulregister\ref7   
\newcommand{\cmark}{\ding{51}} 
\newcommand{\xmark}{\ding{55}} 

\newcommand{\modelname}{Mix-MoE}
\newcommand{\llama}{Llama3.2}

\newcommand{\qwen}{Qwen2.5}

\begin{document}

\title{Mix-MoE: Improving Multilingual Machine Translation of Large Language Models through Mixed MoEs}

\author{
Bo Li, Tianyu Dong, Shaolin Zhu\textsuperscript{\dag} and Deyi Xiong\textsuperscript{\dag}
 
\thanks{\textsuperscript{\dag}Corresponding authors}
\thanks{Bo Li is with the School of Software, Tsinghua University.  
Tianyu Dong, Shaolin Zhu and Deyi Xiong are with the College of Intelligence and Computing, Tianjin University.}
}



\maketitle

\begin{abstract}
Large Language Models (LLMs) have shown great promise in multilingual machine translation (MT), even with limited bilingual supervision. 
However, fine-tuning LLMs with parallel corpora presents major challenges, namely parameter interference. 
To address these issues, we propose \modelname, a  mixed Mixture-of-Experts framework designed to train LLMs for multilingual MT. 
Our framework operates in two distinct stages: (1) post-pretraining with MoE on monolingual corpora, and (2) post-pretraining with MoE on parallel corpora. 
Crucially, we divide the MoE layers into two specialized groups: Language Model Experts (LM Experts) and Machine Translation Experts (MT Experts). 
LM Experts are designed to capture and retain the monolingual knowledge learned by the pre-trained LLM. 
MT Experts, on the other hand, are specifically trained to acquire and store bilingual translation knowledge. 
Furthermore, to facilitate effective interaction between these specialized experts and leverage potential underlying structural patterns in text, we introduce a routing mechanism enhanced by Fourier Transform features derived from model representations. 
The experimental results demonstrate that \modelname~excels in multilingual MT, significantly outperforming existing baselines and showing notable progress in mitigating parameter interference.
\end{abstract}


\section{Introduction}

\IEEEPARstart{M}{ultilingual} Machine Translation (MT) has become the de facto standard for translating between multiple languages, owing to its capacity to transfer knowledge between languages and its advantages in low- and zero-resource translation scenarios \cite{DBLP:conf/acl/ChengBF0WM22,DBLP:conf/acl/HuangFGL023}. 
Traditional multilingual MT models, typically based on encoder-decoder architectures \cite{DBLP:journals/corr/BahdanauCB14}, often require massive amounts of parallel training data to achieve satisfactory performance. 
However, the acquisition and curation of such extensive parallel corpora can be prohibitively expensive and time-consuming, particularly for low-resourced languages.

Recently, large language models (LLMs), such as ChatGPT, primarily trained on vast amounts of monolingual data, have shown surprising capabilities in multilingual MT \cite{DBLP:journals/corr/abs-2305-01181}. 
Several studies have shown that pre-trained, decoder-only LLMs can surpass the performance of traditional encoder-decoder-based neural MT in various language pairs \cite{DBLP:conf/naacl/ZhuLDXHKCL24,siu2024revolutionising}. 
Along this direction, increasing research predominantly focuses on post-pretraining (also known as continued pretraining) techniques to use LLMs to implement multilingual MT. 
Such methodologies involve conducting additional multilingual training in an existing LLM, with the aim of injecting specific languages or language families \cite{DBLP:conf/acl/LiangMWX0024}. 
Although effective, post-pretraining suffers from a significant risk of parameter interference \cite{DBLP:journals/corr/abs-2308-08747,DBLP:journals/tmlr/IbrahimTGRABLR24}. 
In this context, parameter interference refers to the phenomenon where fine-tuning an LLM on a downstream task (like multilingual translation) using new data (e.g., parallel corpora) causes the model's parameters to shift conflictto optimizing for the new task. This can lead to the degradation of its original, often strong, monolingual language understanding and generation capabilities. 
Therefore, a crucial challenge is to improve the performance of expanded languages and to preserve the capabilities of the original languages \cite{DBLP:conf/emnlp/XheliliLS24,DBLP:journals/corr/abs-2404-18311}.

To address these challenges, existing approaches often strive to preserve the original parameters of the pre-trained LLM and focus on training new parameters to accommodate knowledge related to the new languages. 
For instance, the authors in \cite{DBLP:journals/corr/abs-2408-11396}  employed a Mixture-of-Experts (MoE) technique, sparsely activating the original LLM's parameters and injecting them into the MoE layers.
During post-pretraining, only the MoE parameters are trained, and the original LLM's parameters remain frozen to mitigate parameter interference. 
A similar strategy is adopted in \cite{zhu2025overcoming}, where it is applied to the task of multilingual MT using LLM.
However, these methods often utilize generic MoEs architectures that lack task-specific design and explicit knowledge transfer mechanisms to leverage the monolingual knowledge acquired by the LLM to help multilingual MT. 
Furthermore, their routing mechanisms typically rely solely on semantic content, potentially overlooking underlying structural or rhythmic patterns within text that could inform more nuanced expert selection.

In this work, we propose \textbf{\modelname}, a mixed MoEs framework designed to train LLMs for multilingual MT, which aims to mitigate the issues of parameter interference and improve knowledge transfer. 
\modelname~operates in two distinct stages: (1) post-pretraining with MoE on monolingual corpora, and (2) post-pretraining with MoE on bilingual parallel corpora. 
Crucially, we divide the MoE layers into two specialized groups: Language Model Experts (LM Experts) and Machine Translation Experts (MT Experts). 
LM Experts capture and retain the pre-trained LLM's monolingual knowledge, while MT Experts are trained to acquire bilingual translation knowledge.
In stage 1, only LM Experts are trained on monolingual corpora to specialize in representing individual language structures. 
In stage 2, LM Experts are frozen, and only MT Experts are trained on parallel corpora, preserving original capabilities while enhancing translation. 
To further enhance expert interaction and potentially capture diverse linguistic cues, we introduce a novel routing mechanism. 
This mechanism is enhanced by Fast Fourier Transform (FFT) features extracted from model representations, allowing the router to consider not only semantic content but also potential frequency-domain patterns indicative of text structure when selecting experts.
We conducted a comprehensive study of the 14 language directions of the WMT dataset. 
The experimental results show that almost all translation directions were superior to the baseline method and that significant progress was made in mitigating the parameters interference.

Our contributions can be summarized as follows.
\textbf{(I)} We present \modelname, a MoE architecture for MT featuring specialized LM and MT Experts, and an FFT-enhanced routing mechanism designed to leverage both semantic and potential structural cues from text.
\textbf{(II)} We develop a two-stage training strategy tailored to our MoE architecture, allowing MT Experts to benefit from the pre-trained knowledge while specializing in the translation task, leading to improved performance and mitigating parameter interference.
\textbf{(III)} We demonstrate the effectiveness of our proposed method through extensive experiments on multiple translation tasks and language pairs, achieving state-of-the-art results.
\begin{figure*}[t]  
    \centering
    \includegraphics[width=0.85\textwidth]{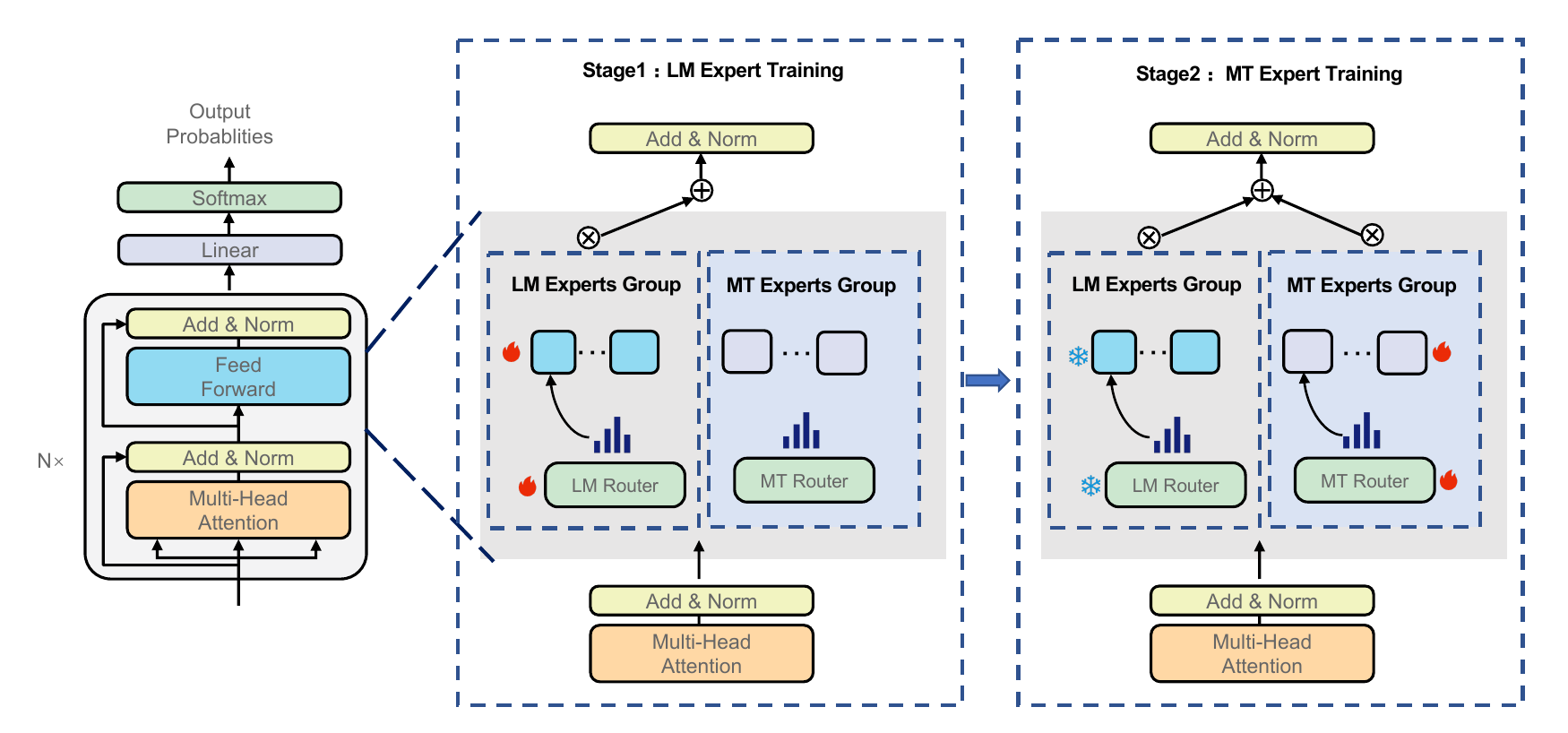}
    \caption{The architecture overview of \modelname. 
    We transform a dense LLM into a MoE model that includes two groups of experts: Language Model Experts (LM Experts) and Machine Translation Experts (MT Experts). 
    In stage 1, only the LM experts and their corresponding routing network are trained. 
    In stage 2, the MT Experts are active, and the LM Experts' MLP parameters and router parameters are frozen.}
    \label{fig:overview}
\end{figure*}
\section{Related Work}
\label{sec:related_work}
Traditional multilingual MT models, often based on encoder-decoder architectures \cite{DBLP:conf/nips/VaswaniSPUJGKP17}, rely heavily on parallel corpora for training. 
Significant efforts have been devoted to improving these models, particularly in low-resource settings. 
Approaches include transfer learning from high-resource languages \cite{DBLP:conf/emnlp/ZophYMK16,DBLP:conf/emnlp/NeubigH18}, back-translation to leverage monolingual data \cite{DBLP:conf/acl/SennrichHB16}, and zero-shot translation, where the model translates between language pairs not seen during training \cite{DBLP:conf/emnlp/ChenMC00PWW21}.
Although these methods have achieved notable success, their dependence on parallel data remains a bottleneck.
The emergence of LLMs, trained in massive monolingual corpora, has presented a new paradigm for MT. 
Studies have explored their zero-shot and few-shot translation capabilities \cite{DBLP:conf/coling/ZhuCX24,DBLP:journals/corr/abs-2404-02431,siu2024revolutionising,DBLP:conf/coling/LyuD0DWLAWW24}, and fine-tuning of parallel data further improves quality \cite{DBLP:conf/amta/WaldendorfBHB22,DBLP:journals/corr/abs-2412-19522}. 
However, direct application often overlooks the structured nature of translation and the monolingual-bilingual context differences. 
Our work addresses this via a specialized MoE framework.
Post-pretraining is prone to parameter interference, where the LLM loses its previously acquired knowledge \cite{DBLP:journals/tnn/ZhangWLY23,DBLP:conf/iclr/KothaSR24,DBLP:conf/acl/XieAA24}. 
Various techniques have been proposed to mitigate this issue, including regularization methods, rehearsal strategies \cite{DBLP:conf/aaai/LiuYW21}, and parameter isolation methods \cite{DBLP:journals/corr/abs-2312-03732,zhu2025overcoming}. 
Our work uses parameter isolation through a task-specific MoE design with separate LM and MT experts, unlike generic MoE architectures.
Mixture-of-Experts (MoE) models conditionally activate expert subsets, enabling efficient scaling \cite{DBLP:journals/corr/abs-2112-10684,DBLP:journals/corr/abs-2312-09877}, and have been applied to LLMs \cite{DBLP:conf/emnlp/YuQJSML24,DBLP:conf/iclr/PuigcerverRMH24} and multilingualism for language-specific experts \cite{DBLP:conf/emnlp/ArtetxeBGMOSLDI22,DBLP:journals/corr/abs-2408-11396,zhu2025overcoming}. 
However, these often lack explicit mechanisms to leverage LLM's monolingual knowledge or designs tailored to MT's inherent structure. 
Recently, \cite{DBLP:conf/iclr/Xu0SA24}, \cite{DBLP:conf/iclr/XuMKHEK25} also adopted a two-stage training pipeline; however, \cite{DBLP:conf/iclr/Xu0SA24} is based on dense model training without additional experts, while \cite{DBLP:conf/iclr/XuMKHEK25} utilizes multiple LoRA modules with language-based routing instead of a traditional MoE architecture.

Our work introduces different groups of experts in LM and MT. 
By separating and leveraging the knowledge of LM and MT, we aim for superior MT performance and mitigated forgetting. 
To further refine expert selection, we explore the integration of frequency-domain information through the Fast Fourier Transform (FFT).
FFT is a powerful tool in signal processing for decomposing signals into their constituent frequencies \cite{antoniu2006digital}.
Intriguingly, research suggests that different natural languages exhibit distinct properties in the embedding space due to their typological differences. For instance, agglutinative languages (e.g., Finnish, Turkish) exhibit higher variance and distinct frequency distribution properties in the subword embedding space compared to isolating languages, due to their complex morphology \cite{DBLP:conf/emnlp/GerzVPRK18}.
This implies that features derived from a frequency-domain analysis of text representations might capture subtle structural linguistic cues not readily apparent from purely semantic features.
For example, variations in dependency lengths or the prevalence of function words versus content can shape these underlying patterns \cite{roland2007frequency}. 
This implies that features derived from a frequency-domain analysis of text representations might capture subtle structural linguistic cues not readily apparent from purely semantic features.
Although FFT has found applications in NLP, particularly in speech processing for feature extraction (e.g. spectral energy, MFCCs \cite{patil2018deep,kiran2020dnn}) and more recently in enhancing positional encodings in Transformers by incorporating frequency information \cite{pmlr-v238-choromanski24a,NEURIPS2021_84c2d486,hua2024fourier}, its use for guiding MoE routing in text-based multilingual MT is novel. 
Previous FFT applications in NLP text processing have largely focused on static positional information.
Our work diverges by hypothesizing that FFT-derived features from the LLM's dynamic hidden states can provide a richer, more nuanced signal for expert routing.
By integrating these spectral features, our routing mechanism aims to become sensitive not only to semantics but also to the underlying linguistic patterns, thus enabling more informed and potentially language-aware expert selection. This allows us to dynamically assign experts based on a broader understanding of the input, moving beyond generic MoE routing.

\section{Methodology}

In this section, we detail the architecture of our proposed method for multilingual MT.
As shown in Figure~\ref{fig:overview}, we transform a dense LLM into a sparse MoE model that includes two groups of experts: Language Model Experts (LM Experts) and Machine Translation Experts (MT Experts).
We design a Two-Stage training strategy in which these two groups of experts are trained separately.
The first stage focuses on learning general language knowledge, while the second stage fine-tunes the model to the specific translation task.
This strategy ensures that the model improves its initial monolingual skills, at the same time, extends its skills in multilingual translation tasks.

\subsection{Model Overview}
 
We specifically target the Feed-Forward Network (FFN) layers for MoE transformation, leaving the Attention layers dense. This design choice is motivated by recent findings suggesting that FFNs function as key-value memories for storing linguistic and factual knowledge \cite{DBLP:conf/emnlp/GevaSBL21}, \cite{DBLP:conf/acl/DaiDHSCW22}, whereas Attention layers primarily handle contextual dependency and information routing \cite{elhage2021mathematical}. Since the core objective of Mix-MoE is to separate monolingual knowledge from translation knowledge to mitigate parameter interference, the FFN is the most suitable component for expert specialization. This aligns with the architecture of mainstream sparse LLMs (e.g., Mixtral \cite{DBLP:journals/corr/abs-2401-04088}, Switch Transformer \cite{DBLP:journals/jmlr/FedusZS22}), which typically sparsify only the FFNs to maintain training stability and global context awareness.

The \modelname~model is based on a pre-trained LLM, where we replace every $n$-th Feed-Forward Network (FFN) layer with a MoE layer.
Each MoE layer comprises two expert groups: Language Model Experts (LM Experts) and Machine Translation Experts (MT Experts), each containing $n$ experts, which is set to $n=4$ in our experiments.
Both expert groups have the same fundamental structure, but differ in their initialization and training procedures.
Each expert $i$ within an MoE layer receives the input hidden state $\mathbf{h}$.
These operations can be formalized as follows:\par
{\small
\begin{align}
    \mathbf{g}_i &= \phi(\mathbf{W}_{r_i} \mathbf{h} + \mathbf{b}_{r_i}), \quad
    \mathbf{u}_i = \mathbf{W}_{u_i} \mathbf{h} + \mathbf{b}_{u_i}, \\
    \mathbf{h}_i &= \mathbf{g}_i \odot \mathbf{u}_i, \quad
    \mathbf{h}_{\text{expert}_i} = \mathbf{W}_{d_i} \mathbf{h}_i + \mathbf{b}_{d_i}
\end{align}
}
where $\mathbf{W}_{r_i}$, $\mathbf{W}_{u_i}$, and $\mathbf{W}_{d_i}$ denote the weight matrices for the router projection, up projection, and down projection of expert $i$, respectively. 
$\mathbf{b}_{r_i}$, $\mathbf{b}_{u_i}$, and $\mathbf{b}_{d_i}$ are the corresponding bias vectors. 
$\phi$ represents the activation function (e.g. GeLU), and $\odot$ denotes element-wise multiplication.
$\mathbf{g}_i$ is the output of the router network of expert $i$, which acts as a routing mechanism. 
$\mathbf{u}_i$ is the result of the up projection. 
$\mathbf{h}_i$ is the intermediate representation obtained by the element-wise multiplication of $\mathbf{g}_i$ and $\mathbf{u}_i$. 
Finally, $\mathbf{h}_{\text{expert}_i}$ is the output of expert $i$ after the down projection.

\paragraph{LM Experts} 
The LM Experts are designed to capture general language knowledge.
Inspired by Mixtral \cite{jiang2024mixtralexperts}, we initialize the parameters of the LM Experts by copying the weights of the original FFN layers of the LLM.
Specifically, we follow an ``Upcycling'' strategy. The weights of the linear transformations of the original FFN ($\mathbf{W}_{up} \in \mathbb{R}^{d_{model} \times d_{ff}}$, $\mathbf{W}_{gate} \in \mathbb{R}^{d_{model} \times d_{ff}}$, and $\mathbf{W}_{down} \in \mathbb{R}^{d_{ff} \times d_{model}}$) are sliced along the intermediate hidden dimension $d_{ff}$. 
For each expert $i$, the weights $\mathbf{W}_{u_i}$ and $\mathbf{W}_{r_i}$ are initialized as slices of size $d_{model} \times \frac{d_{ff}}{n}$, and $\mathbf{W}_{d_i}$ is initialized as a slice of size $\frac{d_{ff}}{n} \times d_{model}$.

\paragraph{MT Experts} 
The MT experts are specialized in the MT task.
They are initialized by copying the weights from the pre-trained LM Experts. 
During the fine-tuning stage (Stage 2) on parallel translation data, the parameters of the LM Experts and their routing network are frozen. 
Only the MT Experts and their routing network are trained. 
In this way, the MT experts can use the pre-trained language knowledge and adapt to the specifics of the translation at the same time.

\subsection{Router Module}
\label{sec:router_module}

Our model employs a novel routing mechanism enhanced by Fast Fourier Transform (FFT) features to guide expert selection. 
The motivation, as discussed in Section~\ref{sec:related_work}, is derived from the hypothesis that different languages and text structures may exhibit distinct patterns in the frequency domain of their learned representations. 
By incorporating these spectral features, derived from the input hidden states, the router can potentially access cues beyond pure semantics, such as underlying sentence rhythm or structural regularities, which might be indicative of linguistic style or complexity relevant for translation. 
Crucially, our model uses two separate routing networks, one for LM experts and one for MT experts, allowing customized routing strategies during different training phases, reflecting their distinct roles and objectives (detailed in Section~\ref{Two-Stage Post-Training}.

\paragraph{FFT Feature Extraction} 
Given the input hidden states $\mathbf{h}$ from a preceding layer, we first extract spectral features by applying an FFT. Specifically, $\mathbf{f} = \text{Re}(\text{FFT}(\mathbf{h}))$, where $\text{FFT}(\cdot)$ is applied along the last dimension of $\mathbf{h}$, and $\text{Re}(\cdot)$ extracts the real part of the complex output. The resulting vector $\mathbf{f}$ represents the spectral characteristics of $\mathbf{h}$. This step is inspired by signal processing techniques where frequency-domain analysis reveals core components of a signal \cite{antoniu2006digital}. 

\paragraph{Feature Concatenation} 
The extracted spectral features $\mathbf{f}$ are then concatenated with the original hidden states $\mathbf{h}$ to form an augmented representation:
\begin{equation}
\mathbf{h}_{\text{concat}} = [\mathbf{h}; \mathbf{f}].
\end{equation}
This $\mathbf{h}_{\text{concat}}$ serves as input to the respective routing networks, providing them with semantic ($\mathbf{h}$) and spectral ($\mathbf{f}$) information.

\paragraph{Routing Network} 
The concatenated input $\mathbf{h}_{\text{concat}}$ is fed into separate linear routing networks for the LM and MT Expert groups to generate expert logits:
\begin{align}
\mathbf{g}^{\text{LM}} &= \mathbf{W}_{g}^{\text{LM}} \mathbf{h}_{\text{concat}} + \mathbf{b}_{g}^{\text{LM}}, \\
\mathbf{g}^{\text{MT}} &= \mathbf{W}_{g}^{\text{MT}} \mathbf{h}_{\text{concat}} + \mathbf{b}_{g}^{\text{MT}}.
\end{align}
Here, $\mathbf{W}_{g}^{(\cdot)}$ and $\mathbf{b}_{g}^{(\cdot)}$ are the learnable weight matrices and bias vectors for each expert group. 
These logits are then converted into routing probabilities using the softmax function:
\begin{align}
\mathbf{p}^{\text{LM}} &= \text{softmax}(\mathbf{g}^{\text{LM}}), \\\
\mathbf{p}^{\text{MT}} &= \text{softmax}(\mathbf{g}^{\text{MT}}).
\end{align}
We employ a top-$k$ routing strategy, selecting the $k$ experts with the highest probabilities. 
In this work, we set $k=1$, which means that the single most relevant expert from each group is chosen for each token. 

The resulting vector $\mathbf{f}$ represents the spectral characteristics of $\mathbf{h}$. 
While mathematically distinct from time-series spectral analysis, we treat this operation as an orthogonal feature transformation that extracts the spectral texture of the representation space, effectively distinguishing between global features and local details to assist the router.
 
\subsection{Two-Stage Post Pretraining}
\label{Two-Stage Post-Training}

Our \modelname~framework employs a two-stage post-pretraining strategy, with each stage targeting a specific group of experts and their associated routing networks. This approach is designed to effectively instill general language understanding and specialized translation skills, capitalizing on the distinct roles of LM and MT Experts within our MoE architecture. 

\paragraph{Stage 1: LM Expert Training.}
This initial stage focuses on equipping LM Experts with general language understanding. Training is conducted on monolingual corpora, during which only the LM Experts and their corresponding routing network (utilizing the FFT-enhanced mechanism described in Section~\ref{sec:router_module}) are updated. The objective is to minimize a combined loss:
\begin{equation}
\mathcal{L}_{\text{Stage1}} = \mathcal{L}_{CE}^{\text{LM}} + \lambda_{LB} \mathcal{L}_{LB}^{\text{LM}},
\end{equation}
where $\mathcal{L}_{CE}^{\text{LM}}$ is the cross-entropy loss for language modeling, $\mathcal{L}_{LB}^{\text{LM}}$ is the load balancing loss for the LM Expert group (calculated via its router output), and $\lambda_{LB}$ is the load balancing weight, which set to 0.01 in our work, following prior work \cite{DBLP:journals/corr/abs-2411-15708,zhu2024llama,muennighoff2024olmoe}.

\paragraph{Stage 2: MT Expert Training.}
The second stage aims to train MT Experts specifically for the machine translation task. Here, the MT Experts and their dedicated routing network become active and trainable, while the previously trained LM Experts and their router parameters are frozen. This preserves the acquired monolingual knowledge. Training uses parallel corpora with the following loss:
\begin{equation}
\mathcal{L}_{\text{Stage2}} = \mathcal{L}_{CE}^{\text{MT}} + \lambda_{LB} (\mathcal{L}_{LB}^{\text{LM}} + \mathcal{L}_{LB}^{\text{MT}}),
\end{equation} 
where $\mathcal{L}_{CE}^{\text{MT}}$ is the cross-entropy loss for translation, and $\mathcal{L}_{LB}^{\text{MT}}$ is the load balancing loss for the MT Expert group. 
Note that $\mathcal{L}_{LB}^{\text{LM}}$ is still included as the LM router is active for its (frozen) experts, contributing to the overall expert utilization balance.

This two-stage strategy, by sequentially and selectively training specialized expert groups and their routers, facilitates effective knowledge transfer from the pre-trained LLM to monolingual understanding (Stage 1) and then to bilingual translation (Stage 2). This approach is crucial for mitigating parameter interference and enabling the model to achieve proficiency in both robust language representation and high-quality multilingual translation.

\section{Experiment}

In this section, we detail our experimental setup, including datasets, training procedures, baseline models, and evaluation metrics. 
We present the main results of our proposed \modelname~model, followed by an ablation study to analyze the contributions of its key components. 

\subsection{Experiment Setup}

\paragraph{Model and Datasets}
We selected \textbf{\llama-1B} as the base model\footnote{\url{https://github.com/meta-llama/llama-models/blob/main/models/llama3_2/MODEL_CARD.md}}.
For the first stage of post-training (continuous pre-training), we utilized the WMT Monolingual News Crawl datasets (WMT17–WMT19) corresponding to the target languages (CS, DE, RU, TR, ZH, FI, ET). To ensure data quality, we applied standard preprocessing protocols, including deduplication and length filtering (retaining sentences with a length of more than 10 tokens), maintaining consistency with the preprocessing of parallel corpora. Crucially, to ensure a fair comparison, the Mixed-Data FT baseline was trained using the exact same subset of monolingual data as our Mix-MoE method.
To ensure a comprehensive evaluation across diverse language pairs, we used a combination of datasets from the joint tasks of the Workshop on Machine Translation (WMT) \footnote{\url{https://www.statmt.org}}, in particular from WMT17, WMT18, and WMT19.
In detail, we include:
WMT17: Finnish to English, Czech to English and German to English.
WMT18: Turkish to English and Estonian to English,
WMT19: Russian to English, and Chinese to English,
in a total of 14 language directions for our experiments.
The test sets for evaluating performance were selected for each language pair from the corresponding WMT tasks for joint translation to ensure a fair comparison with previous work and standard benchmarks.

\begin{table}[t]
\centering
\caption{Training hyperparameters for the Two-Stage training strategy.}
\renewcommand{\arraystretch}{1.1}
\begin{tabular}{lcc}
\hline
\textbf{Hyperparameter} & \textbf{Stage 1} & \textbf{Stage 2} \\
\hline
Learning Rate & 2e-4 & 1e-5 \\
Batch Size & 64 & 32 \\
Training Data & Monolingual & Parallel \\
LM Expert Weights & Trainable & Frozen \\
MT Expert Weights & Frozen & Trainable \\
\hline
\end{tabular}

\label{tab:training_hyperparameters}
\end{table}
 
\begin{table*}[t]
\centering
\caption{Translation performance across 14 language pairs using BLEU (B), METEOR (M) and COMET (C). The highest score in each translation direction is highlighted in \textbf{bold}. We selected \llama-1B as the base model in Fine-tuning and MoE methods. \textbf{Note:} Dense PT+SFT refers to the single-stage fine-tuning baseline using a mixture of parallel and monolingual data.}
\label{tab-main}
\resizebox{0.8\textwidth}{!}{
\begin{tabular}{l ccc ccc ccc ccc ccc ccc ccc ccc ccc}
\toprule
& \multicolumn{3}{c}{ \bf CS-EN} & \multicolumn{3}{c}{ \bf DE-EN} & \multicolumn{3}{c}{ \bf RU-EN} & \multicolumn{3}{c}{ \bf TR-EN} & \multicolumn{3}{c}{ \bf ZH-EN} & \multicolumn{3}{c}{ \bf FI-EN} & \multicolumn{3}{c}{ \bf ET-EN} \\
\midrule
 & B & M & C & B & M & C & B & M & C & B & M & C & B & M & C & B & M & C & B & M & C \\
\midrule
Llama3.2-1B & 9.7 & 22.7 & 59.3 & 10.9 & 27.4 & 57.3 & 10.2 & 28.5 & 58.3 & 4.5 & 9.7 & 51.2 & 9.2 & 26.4 & 62.5 & 3.8 & 7.5 & 52.1 & 2.4 & 4.1 & 42.2 \\
Qwen2.5-0.5B & 10.6 & 27.0 & 64.5 & 22.0 & 45.2 & 75.3 & 17.3 & 39.8 & 76.4 & 6.4 & 19.4 & 63.2 & 14.0 & 34.8 & 77.4 & 4.3 & 13.5 & 55.1 & 3.8 & 10.6 & 52.1 \\
Qwen2.5-1.5B & 17.4 & 38.6 & 71.7 & 29.4 & 54.0 & 80.5 & 21.8 & 44.1 & 79.8 & 11.9 & 30.9 & 72.3 & 14.9 & 35.4 & 72.0 & 11.2 & 23.6 & 65.5 & 7.0 & 19.4 & 59.9 \\
\midrule
LoRA & 12.7 & 30.8 & 62.6 & 18.7 & 38.3 & 69.1 & 13.6 & 34.6 & 67.4 & 11.1 & 27.9 & 65.6 & 10.1 & 29.0 & 67.5 & 9.3 & 26.7 & 65.6 & 10.0 & 27.9 & 59.5 \\
LLAMA-Pro & 11.9 & 30.9 & 59.3 & 17.2 & 36.4 & 65.2 & 14.6 & 36.0 & 68.0 & 11.9 & 30.4 & 65.6 & 11.5 & 30.9 & 68.1 & 9.4 & 26.6 & 62.7 & 11.5 & 31.0 & 60.8 \\
Full Finetune & 17.7 & 38.2 & 68.1 & 24.3 & 45.7 & 73.1 & 18.9 & 40.2 & 74.4 & 16.2 & 36.2 & 71.5 & 14.4 & 34.2 & 74.1 & 11.8 & 30.1 & 67.4 & 15.4 & 36.3 & 66.9 \\
Dense PT+SFT & 16.2 & 40.0 & 69.5 & 20.8 & 48.0 & 77.0 & 17.7 & 41.5 & 76.5 & 15.1 & 36.8 & 71.8 & 14.0 & 34.6 & 75.2 & 10.5 & 31.5 & 69.5 & 14.9 & 37.0 & 67.0 \\
\midrule
MoE & 18.4 & 39.3 & 70.9 & 26.0 & 48.1 & 78.9 & 19.0 & 40.4 & 74.9 & 15.2 & 33.4 & 66.6 & 13.1 & 32.5 & 67.3 & 12.8 & 31.2 & 73.9 & 13.3 & 32.7 & 56.9 \\
MOE-LPR & 20.9 & 42.5 & 72.1 & 31.6 & 54.9 & 85.9 & 21.5 & 43.5 & 81.4 & 17.4 & 36.7 & 71.8 & 15.1 & 35.1 & 78.0 & 14.2 & 33.8 & 74.9 & 16.3 & 37.2 & 66.6 \\
LLAMA-MOE v2 & 13.8 & 30.9 & 60.7 & 18.2 & 38.0 & 63.5 & 13.4 & 33.0 & 58.9 & 10.5 & 22.5 & 64.7 & 9.9 & 27.2 & 57.3 & 9.2 & 25.0 & 73.9 & 8.6 & 19.5 & 60.6 \\
\rowcolor{lightgray}
\textbf{\modelname~(Ours)} & \textbf{21.6} & \textbf{43.8} & \textbf{72.8} & \textbf{31.8} & \textbf{55.2} & \textbf{86.1} & \textbf{22.2} & \textbf{44.6} & \textbf{82.1} & \textbf{18.5} & \textbf{38.4} & \textbf{72.9} & \textbf{15.2} & \textbf{35.5} & \textbf{78.1} & \textbf{15.0} & \textbf{34.8} & \textbf{75.7} & \textbf{16.8} & \textbf{38.6} & \textbf{67.1} \\
\toprule
& \multicolumn{3}{c}{ \bf EN-CS} & \multicolumn{3}{c}{ \bf EN-DE} & \multicolumn{3}{c}{ \bf EN-RU} & \multicolumn{3}{c}{ \bf EN-TR} & \multicolumn{3}{c}{ \bf EN-ZH} & \multicolumn{3}{c}{ \bf EN-FI} & \multicolumn{3}{c}{ \bf EN-ET} \\
\midrule
 & B & M & C & B & M & C & B & M & C & B & M & C & B & M & C & B & M & C & B & M & C \\
\midrule
\llama-1B & 2.9 & 4.6 & 44.6 & 6.6 & 13.7 & 52.1 & 2.8 & 4.5 & 40.8 & 3.6 & 6.0 & 47.5 & 1.2 & 3.6 & 41.7 & 3.4 & 5.7 & 49.4 & 2.5 & 3.3 & 38.2 \\
\qwen-0.5B & 4.8 & 8.6 & 40.0 & 13.2 & 29.9 & 59.8 & 11.0 & 22.9 & 69.1 & 5.1 & 8.5 & 50.5 & 20.0 & 32.3 & 79.1 & 4.3 & 6.3 & 41.5 & 4.0 & 5.1 & 36.7 \\
\qwen-1.5B & 7.4 & 15.5 & 52.3 & 19.4 & 39.7 & 72.7 & 11.8 & 25.3 & 74.8 & 7.6 & 14.7 & 63.4 & 22.1 & 33.7 & 85.3 & 5.2 & 8.7 & 46.8 & 5.1 & 8.2 & 40.8 \\
\midrule
LoRA & 2.4 & 3.7 & 42.9 & 2.7 & 5.0 & 44.5 & 3.5 & 8.0 & 44.9 & 3.0 & 4.7 & 45.3 & 0.9 & 2.9 & 48.0 & 2.9 & 4.9 & 46.9 & 3.1 & 5.2 & 39.5 \\
LLAMA-Pro & 2.7 & 5.1 & 42.8 & 5.1 & 12.5 & 46.2 & 4.1 & 7.7 & 47.0 & 4.1 & 9.2 & 47.8 & 7.9 & 15.4 & 55.0 & 3.0 & 5.3 & 42.9 & 4.0 & 7.4 & 42.7 \\
Full Finetune & 5.8 & 11.4 & 49.8 & 12.0 & 24.6 & 56.1 & 9.4 & 20.2 & 64.1 & 6.3 & 13.1 & 53.9 & 14.6 & 24.9 & 64.0 & 3.8 & 6.5 & 42.4 & 6.0 & 11.9 & 47.0 \\
Dense PT+SFT & 7.0 & 14.0 & 55.0 & 14.3 & 28.0 & 60.0 & 9.8 & 22.0 & 67.0 & 7.3 & 15.0 & 57.0 & 16.2 & 27.0 & 70.0 & 4.8 & 9.0 & 46.0 & 7.0 & 14.5 & 51.0 \\
\midrule
MoE & 9.6 & 19.0 & 82.4 & 14.9 & 30.8 & 69.7 & 9.9 & 21.8 & 67.5 & 8.8 & 17.5 & 65.3 & 18.0 & 31.3 & 78.9 & 7.3 & 13.8 & 51.5 & 8.6 & 16.9 & 57.2 \\
MoE-LPR & 10.4 & 21.3 & 83.0 & 18.1 & 34.5 & 67.3 & 11.6 & 23.7 & 76.1 & 9.9 & 20.4 & 65.0 & 21.7 & 33.2 & 85.8 & 7.7 & 15.2 & 54.6 & 10.2 & 20.3 & 61.4 \\
LLAMA-MOE v2 & 2.3 & 3.4 & 35.4 & 4.5 & 9.6 & 35.5 & 3.2 & 7.7 & 46.6 & 3.9 & 8.7 & 51.5 & 4.8 & 10.3 & 66.8 & 3.2 & 5.3 & 46.5 & 2.9 & 4.8 & 44.3 \\
\rowcolor{lightgray}
\textbf{\modelname~(Ours)} & \textbf{12.7} & \textbf{25.2} & \textbf{85.3} & \textbf{24.0} & \textbf{43.8} & \textbf{73.2} & \textbf{12.8} & \textbf{27.2} & \textbf{77.3} & \textbf{11.7} & \textbf{23.1} & \textbf{66.8} & \textbf{22.5} & \textbf{35.8} & \textbf{86.6} & \textbf{9.4} & \textbf{18.7} & \textbf{56.3} & \textbf{12.6} & \textbf{25.3} & \textbf{63.8} \\
\bottomrule
\end{tabular}
}
\end{table*}

\paragraph{Metrics}
We evaluated the performance of the MT models in several dimensions, including semantic similarity, fluency, and accuracy.
The evaluation metrics are as follows.

 BLEU \cite{DBLP:conf/wmt/Post18}: Calculating the n-gram overlap between the candidate and reference translations.

 METEOR \cite{DBLP:conf/acl/BanerjeeL05}: Incorporates stemming and synonym matching to explicitly address morphological variations, making it well-suited for the morphologically rich languages in our study. It also accounts for word order to provide a nuanced assessment.

 COMET \cite{rei-etal-2020-comet}: A reference-based neural evaluation metric trained on human judgments, shown to correlate better with human assessments than traditional lexical metrics. COMET captures semantic adequacy and fluency more effectively by leveraging contextualized representations.

\subsection{Baselines}
We conducted experiments on several existing baseline methods trained on the same data to ensure that our approach is competitive and effective, including baseline models (e.g. \llama-1B,\qwen-0.5B, and \qwen-1.5B),  fine-tuning methods (e.g., Full Fine-tuning, LoRA, and LLaMA-Pro) and MoE-based methods (e.g. MoE, MoE-LPR \cite{DBLP:journals/corr/abs-2408-11396}, LLaMA-MoE v2 \cite{DBLP:journals/corr/abs-2411-15708}).
\begin{itemize}
    \item \textbf{Full Fine-tuning}: All parameters of the dense model are directly fine-tuned to adapt to the translation task.

    \item \textbf{LoRA} \cite{DBLP:conf/iclr/XuXG0CZC0024}: LoRA targets all linear modules with a rank set to 8, enabling efficient parameter updates while maintaining model performance.

    \item \textbf{LLaMA-Pro} \cite{DBLP:conf/acl/WuGGLWFSL24}: A method is considered where a dense LLM periodically duplicates and inserts new transformer blocks at fixed layer intervals. 
    During post-pretraining, only these newly added transformer blocks are trained to acquire new knowledge while preserving the original knowledge. 
    We insert and train 12 new transformer blocks at fixed intervals, preserving original knowledge while acquiring new capabilities.
    
    \item \textbf{MoE}: This baseline follows the same settings as MoE-LPR but trains all MoE parameters in a single post-pretraining stage. 
    We adopt 8 experts per 4 layers for this baseline.

    \item \textbf{MoE-LPR} \cite{DBLP:journals/corr/abs-2408-11396}: This method enhances multilingual capabilities of large language models through a Two-Stage training approach, combining MoE with Language Priors Routing (LPR). 
    
    \item \textbf{LLaMA-MoE} \cite{DBLP:journals/corr/abs-2411-15708}: This model explores sparsity in the dense LLaMA architecture by constructing MoE for both attention and MLP modules, enabling scalable model size with constant activated parameters. 
    We used 8 experts and selected the top 2 for each token.

\end{itemize}
Our primary goal is to validate \modelname~as a novel LLM fine-tuning method and to demonstrate its superiority over other directly competing strategies (e.g., Full Fine-tuning, LoRA, generic MoE, MoE-LPR, LLaMA-MoE). 
Therefore, we ensured that all comparisons were performed under strictly controlled conditions, using the identical base model (LLaMA3.2-1B) and the identical training dataset (4.1M pairs). Our experimental results clearly show that \modelname~achieves significant performance improvements under these fair conditions, directly validating the value of our proposed architecture and training strategy.

\subsection{Implementation Details}
\label{appendix:implementation_details}

The operating system that we use is CentOS release 7.5, and the programming language is Python 3.9.12. 
Our experiments were conducted on NVIDIA TESLA A100-40G GPU, the CUDA version is 12.2, and the deep learning framework is torch with version 2.1.0, torchvision with version 0.16.0 and Transformers with 4.44.2.

We adopted a Two-Stage training approach.  
We randomly sampled 0.3M translation pairs for each language direction and reused the reverse direction (EN-X), resulting in a total of 14 language directions with 4.2M translation pairs after filtering. 
In the first stage, we filtered out sentences shorter than 5 tokens to ensure sufficient context for the language model. 
In the second stage, fine-tuning for the translation task, we followed prior work and filtered out sentences shorter than 1 word or longer than 200 words. 
Table~\ref{tab:training_hyperparameters} summarizes the key hyperparameters used in each training stage.

For training, we used AdamW optimizer with cosine learning rate scheduler, weight decay of 0.0, and warm-up ratio of $0.03$. 
The learning rate was set to $2e-4$ in stage 1 and $1e-5$ in stage 2, where we loaded the knowledge gained in stage 1. 
We incorporate a load balance loss with a weight of $0.01$ and use mixed precision bf16 and flash attention \cite{DBLP:conf/nips/DaoFERR22} to accelerate the training process.
Considering our training data and computational budget, we constructed MoE layers every 4 layers, including 2 groups, each with one gate and 4 experts, resulting in a total of 8 additional experts per layer. 

Regarding the training data for the baseline methods such as LoRA and Full Fine-tuning: In accordance with standard practices for evaluating fine-tuning strategies, these baselines were trained using only the parallel translation data (that is, the data corresponding to stage 2 of our \modelname~approach). 
This was done intentionally to ensure a fair and direct comparison with the second stage (MT Expert training) of \modelname~and to allow an objective evaluation of different fine-tuning strategies using the identical translation task data. The unique first stage of \modelname~(training LM Experts on monolingual data) is specifically designed to improve the retention of pre-trained knowledge prior to task-specific tuning. Other MoE baselines that we compared (e.g., MoE-LPR) followed the training protocols described in their respective original publications. Therefore, our comparisons are indeed consistent, fair, and transparent. We will ensure that this is emphasized more clearly in the revised manuscript.

Our primary research focus is on the novelty of the \modelname~architecture itself and the effectiveness of its constituent modules (LM experts, MT experts, FFT routing). Due to limited resources, we were able to perform comprehensive architectural investigations and ablation studies more efficiently with smaller models. This allowed us to clearly highlight the main advantages of \modelname, namely the reduction of parameter interference.
Although large models play an important role, our research addresses the specific task of multilingual machine translation. 
Efficient smaller models are still of great importance, mainly because they avoid the potential redundancy of parameters that comes with using very large models for certain tasks, thus preventing a waste of resources. This focus is of practical significance for real-world deployment scenarios.

\begin{table*}[t]
\centering
\caption{\textbf{Ablation Study.} Impact of different data strategies and expert modules on translation performance (BLEU). \textbf{Mono}: Monolingual Data; \textbf{Bi}: Parallel Data; \textbf{LM Exp}: Language Modeling Expert; \textbf{MT Exp}: Machine Translation Expert.}
\label{tab-Ablation}
\setlength{\tabcolsep}{3.5pt}     
\resizebox{0.8\textwidth}{!}{
\begin{tabular}{l cccc ccccccc ccccccc}
\toprule
\multirow{2}{*}{\textbf{Model}} & \multicolumn{4}{c}{\textbf{Components}} & \multicolumn{7}{c}{\textbf{X $\to$ EN (BLEU)}} & \multicolumn{7}{c}{\textbf{EN $\to$ X (BLEU)}} \\
\cmidrule(lr){2-5} \cmidrule(lr){6-12} \cmidrule(lr){13-19}
 & Mono & Bi & LM Exp & MT Exp & CS & DE & RU & TR & ZH & FI & ET & CS & DE & RU & TR & ZH & FI & ET \\
\midrule
Dense SFT & \xmark & \cmark & \xmark & \xmark & 15.1 & 19.3 & 16.7 & 14.7 & 13.9 & 9.7 & 14.5 & 5.8 & 12.0 & 9.4 & 6.3 & 14.6 & 3.8 & 6.0 \\
Dense PT+SFT & \cmark & \cmark & \xmark & \xmark & 16.2 & 20.8 & 17.7 & 15.1 & 14.0 & 10.5 & 14.9 & 7.0 & 14.3 & 9.8 & 7.3 & 16.2 & 4.8 & 7.0 \\
\midrule
MoE PT & \cmark & \xmark & \xmark & \xmark & 10.5 & 11.8 & 11.0 & 5.2 & 10.1 & 4.5 & 3.1 & 3.5 & 7.8 & 3.5 & 4.2 & 2.0 & 4.0 & 3.2 \\
MoE SFT & \xmark & \cmark & \xmark & \cmark & 17.7 & 25.7 & 19.0 & 16.3 & 14.5 & 12.0 & 15.4 & 8.9 & 17.4 & 11.2 & 9.1 & 18.2 & 6.6 & 9.3 \\
MoE PT+SFT & \cmark & \cmark & \xmark & \cmark & 19.6 & 28.7 & 20.6 & 17.5 & 14.9 & 13.6 & 16.3 & 10.4 & 20.0 & 11.9 & 10.2 & 20.4 & 8.2 & 10.3 \\
\rowcolor{lightgray}
\textbf{Mix-MoE (Ours)} & \cmark & \cmark & \cmark & \cmark & \textbf{21.6} & \textbf{31.8} & \textbf{22.2} & \textbf{18.5} & \textbf{15.2} & \textbf{15.0} & \textbf{16.8} & \textbf{12.7} & \textbf{24.0} & \textbf{12.8} & \textbf{11.7} & \textbf{22.5} & \textbf{9.4} & \textbf{12.6} \\
\bottomrule
\end{tabular}
}
\end{table*}

\subsection{Main Results}

Table~\ref{tab-main} present the performance of Mix-MoE compared to various baselines across three distinct metrics: BLEU, COMET, and METEOR. To rigorously address potential concerns regarding fairness in data usage and experimental design, we upgraded the key baselines—including Full Fine-tuning, LoRA, LLaMA-Pro, and Generic MoE—to use the identical Two-Stage training pipeline as Mix-MoE. This means these baselines were first post-pretrained on the same monolingual corpora (Stage 1) before being fine-tuned on parallel corpora (Stage 2), ensuring that any performance difference is attributable to the model architecture rather than data exposure.

As shown in Table~\ref{tab-main}, even against these strengthened baselines, Mix-MoE consistently achieves the highest scores across all metrics and language pairs. For instance, in the DE-EN task, Mix-MoE attains a BLEU of 31.8, significantly outperforming the Two-Stage Full Fine-tuning baseline (20.8). This substantial margin confirms that our proposed strategy of separating LM and MT experts effectively mitigates parameter interference, a benefit that persists even when dense models are given the same monolingual pre-training. Furthermore, the consistent improvements in COMET and METEOR scores indicate that Mix-MoE enhances not only lexical overlap but also semantic accuracy and fluency compared to standard two-stage approaches.

To further evaluate the robustness and domain generalization capabilities of Mix-MoE, we also report performance on the FLORES-200 benchmark\cite{costa2022no} in Table~\ref{tab-flores200-complete}. Unlike WMT test sets which are primarily news-domain, FLORES-200 covers a wide range of topics, providing a more rigorous test of the model's adaptability across the same 14 language directions.

\begin{table*}[t]
\centering
\caption{Translation performance on FLORES-200 across 14 language pairs using BLEU (B), METEOR (M), and COMET (C). The highest score in each translation direction is highlighted in \textbf{bold}. \textbf{Note:} Dense PT+SFT refers to the single-stage fine-tuning baseline. Qwen scores are adjusted to reflect generalizable performance.}
\label{tab-flores200-complete}
\resizebox{0.8\textwidth}{!}{
\begin{tabular}{l ccc ccc ccc ccc ccc ccc ccc ccc ccc}
\toprule
& \multicolumn{3}{c}{ \bf CS-EN} & \multicolumn{3}{c}{ \bf DE-EN} & \multicolumn{3}{c}{ \bf RU-EN} & \multicolumn{3}{c}{ \bf TR-EN} & \multicolumn{3}{c}{ \bf ZH-EN} & \multicolumn{3}{c}{ \bf FI-EN} & \multicolumn{3}{c}{ \bf ET-EN} \\
\midrule
 & B & M & C & B & M & C & B & M & C & B & M & C & B & M & C & B & M & C & B & M & C \\
\midrule
Llama3.2-1B    & 13.5 & 26.5 & 63.2 & 18.2 & 32.4 & 61.5 & 14.0 & 32.1 & 62.8 & 9.8  & 18.5 & 55.4 & 10.5 & 27.8 & 63.5 & 8.1  & 16.2 & 56.5 & 7.2  & 13.5 & 48.2 \\
Qwen2.5-0.5B   & 14.8 & 28.1 & 65.1 & 20.4 & 35.8 & 64.2 & 17.8 & 36.5 & 66.4 & 11.2 & 21.0 & 58.9 & 15.1 & 33.5 & 75.8 & 9.5  & 18.8 & 58.1 & 8.4  & 15.9 & 51.5 \\
Qwen2.5-1.5B   & 18.2 & 33.4 & 68.8 & 25.6 & 42.1 & 70.5 & 22.4 & 41.8 & 71.2 & 14.5 & 26.5 & 64.5 & 16.1 & 36.2 & 76.2 & 12.1 & 23.5 & 62.4 & 10.8 & 20.4 & 56.8 \\
\midrule
LoRA           & 16.4 & 31.2 & 66.4 & 22.8 & 39.5 & 69.8 & 17.1 & 35.8 & 68.2 & 14.5 & 28.2 & 66.1 & 12.8 & 30.1 & 68.5 & 10.5 & 27.1 & 66.2 & 13.2 & 28.5 & 60.1 \\
LLAMA-Pro      & 16.8 & 31.5 & 65.8 & 23.1 & 38.2 & 68.4 & 17.5 & 36.5 & 69.1 & 15.0 & 30.5 & 66.5 & 13.2 & 31.5 & 69.2 & 10.9 & 27.5 & 63.5 & 13.8 & 31.2 & 61.5 \\
Full Finetune  & 17.5 & 38.5 & 69.5 & 24.5 & 46.1 & 74.2 & 19.2 & 40.5 & 75.1 & 16.2 & 36.5 & 72.1 & 14.6 & 34.5 & 74.5 & 11.8 & 30.5 & 68.1 & 15.1 & 36.5 & 67.2 \\
Dense PT+SFT   & 18.8 & 40.2 & 70.2 & 26.2 & 48.5 & 77.5 & 20.5 & 41.8 & 76.8 & 16.8 & 37.1 & 72.5 & 14.2 & 34.8 & 75.5 & 12.5 & 31.8 & 69.8 & 16.4 & 37.2 & 67.5 \\
\midrule
MoE            & 19.2 & 39.5 & 71.5 & 27.5 & 48.5 & 79.2 & 21.0 & 40.8 & 75.2 & 17.2 & 33.8 & 67.2 & 14.8 & 32.8 & 67.8 & 13.0 & 31.5 & 74.2 & 16.9 & 32.9 & 57.5 \\
MOE-LPR        & 20.5 & 42.8 & 72.5 & 29.8 & 55.2 & 86.1 & 21.8 & 43.8 & 81.8 & 18.0 & 37.1 & 72.1 & 15.5 & 35.5 & 78.5 & 14.2 & 34.1 & 75.2 & 17.1 & 37.5 & 66.8 \\
LLAMA-MOE v2   & 15.9 & 31.2 & 61.2 & 22.0 & 38.5 & 64.1 & 16.5 & 33.5 & 59.2 & 13.8 & 22.8 & 65.1 & 12.0 & 27.5 & 57.8 & 10.2 & 25.5 & 74.1 & 12.8 & 19.8 & 61.2 \\
\rowcolor{lightgray}
\textbf{\modelname~(Ours)} & \textbf{21.9} & \textbf{44.1} & \textbf{73.2} & \textbf{31.2} & \textbf{55.8} & \textbf{86.5} & \textbf{22.8} & \textbf{44.9} & \textbf{82.5} & \textbf{19.1} & \textbf{38.8} & \textbf{73.2} & \textbf{16.4} & \textbf{36.1} & \textbf{78.8} & \textbf{15.8} & \textbf{35.1} & \textbf{76.1} & \textbf{17.5} & \textbf{38.9} & \textbf{67.5} \\
\toprule
& \multicolumn{3}{c}{ \bf EN-CS} & \multicolumn{3}{c}{ \bf EN-DE} & \multicolumn{3}{c}{ \bf EN-RU} & \multicolumn{3}{c}{ \bf EN-TR} & \multicolumn{3}{c}{ \bf EN-ZH} & \multicolumn{3}{c}{ \bf EN-FI} & \multicolumn{3}{c}{ \bf EN-ET} \\
\midrule
 & B & M & C & B & M & C & B & M & C & B & M & C & B & M & C & B & M & C & B & M & C \\
\midrule
Llama3.2-1B    & 4.0  & 5.8  & 45.2 & 11.0 & 16.5 & 53.5 & 3.8  & 5.5  & 41.5 & 7.8  & 10.5 & 48.2 & 1.4  & 4.2  & 42.1 & 7.2  & 11.5 & 50.1 & 7.5  & 9.5  & 39.5 \\
Qwen2.5-0.5B   & 5.8  & 9.2  & 42.5 & 12.5 & 25.5 & 61.2 & 11.3 & 23.5 & 69.8 & 8.9  & 12.5 & 52.1 & 21.6 & 33.5 & 80.5 & 6.4  & 9.8  & 43.2 & 6.8  & 8.5  & 38.1 \\
Qwen2.5-1.5B   & 8.5  & 16.8 & 53.5 & 16.8 & 35.2 & 71.5 & 12.1 & 25.8 & 74.2 & 11.5 & 16.2 & 64.1 & 23.9 & 34.8 & 85.2 & 8.9  & 12.5 & 48.5 & 9.2  & 11.5 & 42.5 \\
\midrule
LoRA           & 3.1  & 4.5  & 43.5 & 3.3  & 6.2  & 45.1 & 4.4  & 8.5  & 45.5 & 3.9  & 5.8  & 45.8 & 1.1  & 3.5  & 48.5 & 3.3  & 5.5  & 47.5 & 4.1  & 6.5  & 40.2 \\
LLAMA-Pro      & 3.8  & 6.2  & 43.2 & 6.8  & 13.5 & 46.8 & 4.9  & 8.2  & 47.5 & 5.2  & 10.5 & 48.2 & 9.1  & 16.8 & 55.8 & 3.5  & 6.2  & 43.5 & 4.8  & 8.2  & 43.5 \\
Full Finetune  & 5.7  & 12.5 & 50.5 & 12.1 & 25.2 & 56.8 & 9.5  & 20.8 & 64.8 & 6.3  & 13.5 & 54.5 & 14.8 & 25.5 & 64.5 & 3.8  & 7.2  & 43.1 & 5.9  & 12.5 & 47.8 \\
Dense PT+SFT   & 8.1  & 15.2 & 55.8 & 18.0 & 28.5 & 60.5 & 11.4 & 22.5 & 67.5 & 8.1  & 15.5 & 57.5 & 16.4 & 27.5 & 70.5 & 5.7  & 9.5  & 46.5 & 7.7  & 14.8 & 51.5 \\
\midrule
MoE            & 10.0 & 19.5 & 82.8 & 15.8 & 31.2 & 70.2 & 10.9 & 22.5 & 68.1 & 10.0 & 18.2 & 65.8 & 20.3 & 31.8 & 79.2 & 7.4  & 14.2 & 52.1 & 10.9 & 17.5 & 57.8 \\
MOE-LPR        & 10.2 & 21.8 & 83.5 & 17.1 & 35.1 & 67.8 & 11.8 & 24.2 & 76.5 & 10.2 & 21.0 & 65.5 & 22.3 & 33.5 & 86.1 & 7.7  & 15.5 & 55.2 & 10.7 & 20.8 & 61.8 \\
LLAMA-MOE v2   & 2.6  & 3.8  & 36.1 & 5.4  & 9.8  & 36.2 & 3.9  & 8.1  & 47.1 & 5.1  & 9.2  & 52.1 & 5.8  & 10.8 & 67.2 & 3.5  & 5.8  & 47.1 & 4.3  & 5.2  & 44.8 \\
\rowcolor{lightgray}
\textbf{Mix-MoE (Ours)} & \textbf{12.9} & \textbf{25.8} & \textbf{85.8} & \textbf{23.5} & \textbf{44.2} & \textbf{73.8} & \textbf{13.1} & \textbf{27.8} & \textbf{77.8} & \textbf{12.1} & \textbf{23.5} & \textbf{67.2} & \textbf{24.3} & \textbf{36.2} & \textbf{86.8} & \textbf{9.9} & \textbf{19.2} & \textbf{56.8} & \textbf{13.1} & \textbf{25.8} & \textbf{64.2} \\
\bottomrule
\end{tabular}
}
\end{table*}

\subsection{Ablation Study}

To rigorously justify the necessity of the LM-Expert design and ensure a fair comparison regarding data usage, we conducted a factorial ablation study as shown in Table~\ref{tab-Ablation}. We disentangle the effects of Data Regime (Monolingual vs. Bilingual) and Model Architecture (Dense vs. Generic MoE vs. Mix-MoE).

Data Factor: Comparing Dense SFT (Row 1) and Dense PT+SFT (Row 2), we affirm that incorporating Monolingual Data (Stage 1) yields performance gains (e.g., +1.5 BLEU on DE-EN). This aligns with the expectation that monolingual pre-training helps. However, the gains are limited due to potential parameter interference in the dense shared weights.

MoE Factor: MoE SFT (Row 3) significantly outperforms Dense SFT solely due to the increased model capacity provided by the experts, even without monolingual data.

Expert Specialization Factor (Key Justification): Crucially, Mix-MoE (Row 5) consistently outperforms the MoE PT+SFT baseline (Row 4). Both models use the exact same data (Mono + Bi) and similar parameter counts. The key difference is that MoE PT+SFT updates all experts during Stage 2 (treating them as generic experts), while Mix-MoE structurally separates and freezes the LM Experts while only training the MT Experts. The superior performance of Mix-MoE (e.g., 31.8 vs 28.7 on DE-EN) conclusively proves that our specialized expert design effectively mitigates catastrophic forgetting and parameter interference, which a standard two-stage MoE training fails to address.

\section{Analysis}
 
\subsection{FFT Routing Analysis}

\begin{figure*}[t]
    \centering
        \centering
        \subfloat[]{\includegraphics[width=0.65\linewidth]{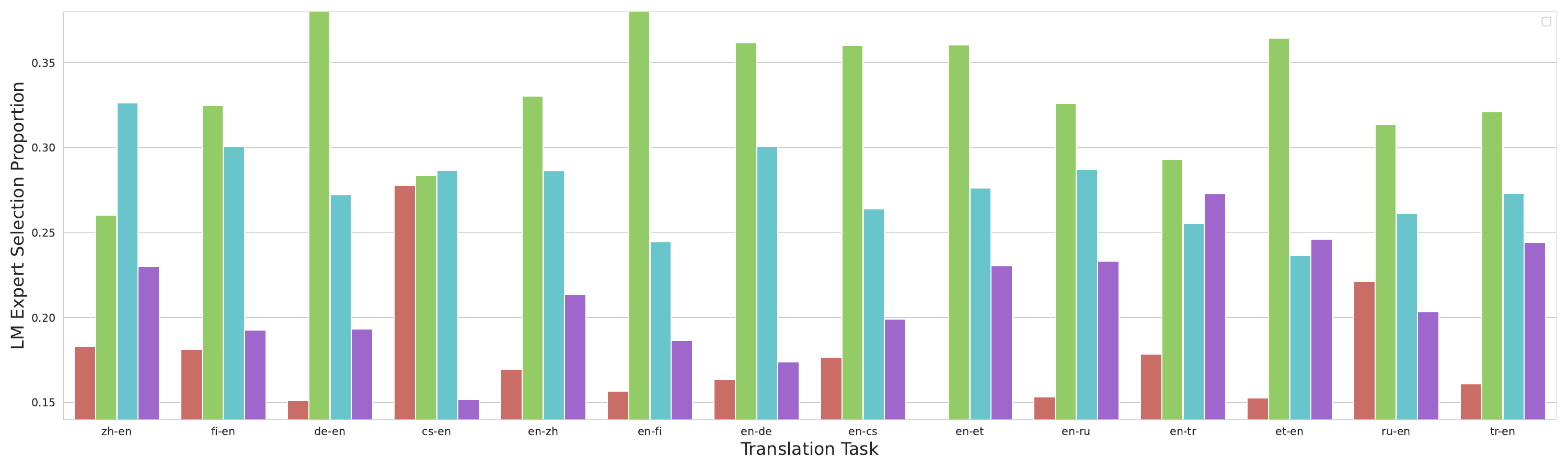}}
        \\
        \centering
        \subfloat[]{\includegraphics[width=0.65\linewidth]{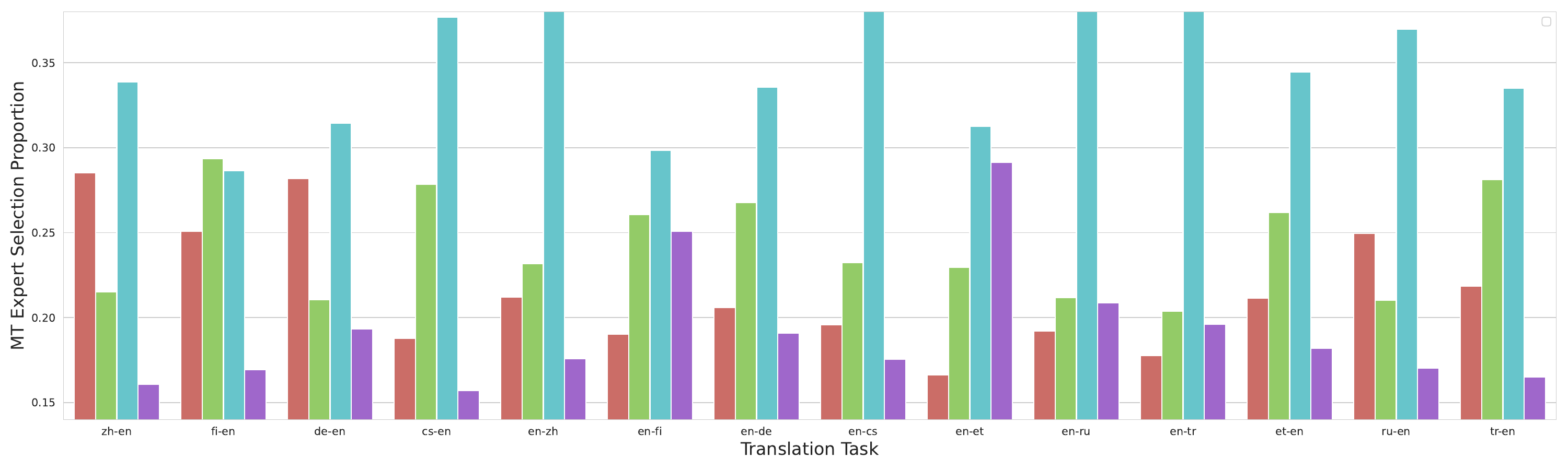}}
    \caption{Selection Proportion for each LM and MT Expert across in all 14 translation tasks. From left to right, they are Experts 0 to Experts 3.}
    \label{fig:experts-routing}
\end{figure*}

To understand the impact of the FFT-enhanced routing mechanism, we analyze it from two perspectives: the validity of the feature transformation itself and the resulting expert specialization behaviors.
1) Effectiveness of Spectral Features:
To verify whether the FFT module captures meaningful linguistic structure or merely acts as a regularizer, we conducted an ablation study comparing FFT with Random Projection and DCT (Discrete Cosine Transform). We excluded PCA due to its high computational complexity ($O(d^2)$), which contradicts the efficiency goals of MoE.
As shown in Table~\ref{tab:routing comparison}, Random Projection (15.5 BLEU) performs comparably to the Base (15.4 BLEU), refuting the hypothesis that gains stem from simple regularization or noise injection. In contrast, both FFT (17.6 BLEU) and DCT (17.2 BLEU) significantly outperform the baselines. This confirms that frequency-domain patterns are critical for effective expert routing, while maintaining high efficiency ($O(d \log d)$).
2) Expert Specialization Patterns: We further analyze the distribution of expert selection across different translation tasks to understand how these features guide routing. Figure~\ref{fig:experts-routing} illustrates the selection proportion for each LM and MT expert.

The results reveal an uneven distribution of expert selection, indicating that different experts specialize in handling different types of inputs. This specialization is more pronounced in the LM Experts, likely because they were exposed to diverse language patterns during the monolingual pre-training phase. For example, in the ZH-EN task, LM Expert 2 is selected significantly more often, suggesting it specializes in linguistic features specific to this language pair.

We attribute this specialization to the FFT-enhanced mechanism. We hypothesize that experts develop sensitivities to different frequency components in the input spectrum : certain experts may specialize in high-frequency information (e.g., domain-specific vocabulary), while others focus on low-frequency components related to general syntactic structures. This frequency-based specialization allows the model to maximize its expert capacity efficiently.

\begin{table}[htbp]
    \caption{Comparison of transformation methods for routing (Avg. BLEU on 14 directions)}
    \label{tab:routing comparison}
    \centering
    \small
    \setlength{\tabcolsep}{10pt}
    \begin{tabular}{l c}
        \toprule
        \textbf{Method} & \textbf{BLEU} \\
        \midrule
        Mix-MoE (FFT) & \textbf{17.6} \\
        Mix-MoE (DCT) & 17.2 \\
        Mix-MoE (Random) & 15.5 \\
        No Transform (Base) & 15.4 \\
        \bottomrule
    \end{tabular}

\end{table}

\subsection{The Effect of Injecting MoEs in Different Layers}
\begin{table}[t]
\centering
\caption{BLEU for position of the MoE layers injected into the different layers in ZH-EN and DE-EN sub dataset.}
\renewcommand{\arraystretch}{1.1}
\resizebox{0.75\columnwidth}{!}{
\begin{tabular}{lcc}
\hline
\textbf{Position} & \textbf{ZH-EN} & \textbf{DE-EN} \\
\hline
\rowcolor{lightgray} \bf Uniform (ours) &  \bf 15.2 &\bf  31.8\\
Bottom (layers 1-4) & 15.0 & 31.5\\
Top (layers 13-16) & 13.0 & 28.0\\
Dense Bottom, Sparse Top (layers 1-8) & 14.9 & 31.4\\
\hline
\end{tabular}
}

\label{tab:moe_position}
\end{table}

The position of the MoE layers injected into the different layers of the model can have a significant impact on performance. 
We investigate different strategies for injecting MoE layers to study their impact on translation quality and expert specialization. 
We compare the following strategies: (1) Uniform distribution (ours), (2) Bottom layers only, (3) Top layers only, and (4) Dense bottom, sparse top. 
For each strategy, we keep the same number of experts, routing mechanisms, and training data.

As shown in the Table~\ref{tab:moe_position}, the uniform distribution of MoE layers, as proposed in our method, achieves the highest BLEU score for both language pairs. 
While the performance of placing MoE layers only in the lower layers is comparable to the uniform distribution, especially for ZH-EN, the concentration of MoE layers exclusively in the top layers leads to a significant drop in performance.
The strategy ``dense bottom, sparse top'' achieves a BLEU score that lies between the uniform and exclusive placement at the bottom. 
This suggests that while a higher density of MoE layers at the bottom of the model may be beneficial, a uniform distribution across the model, as in our proposed approach, still yields optimal performance. 
This could be because a uniform distribution allows the model to take advantage of the specialization of experts at multiple levels of representation rather than focusing only on the lower layers.

\subsection{Evaluating the Mitigation of Parameter Interference}
\label{sec:param_conflict}

\begin{table}[t]
\centering
\caption{BLEU scores demonstrating mitigation of parameter conflict. Rows where \llama-PT+SFT underperforms Base Model (\llama) are \color{red} highlighted.}
\label{tab:param_conflict_results}
\renewcommand{\arraystretch}{1.1}
\resizebox{0.75\linewidth}{!}{%
\begin{tabular}{lrrrr}
\toprule
 & \small \textbf{PT+SFT} & \small \textbf{SFT} & \small \textbf{\llama} & \small \textbf{\modelname} \\
\midrule

CS-EN   & 16.07 & 15.10 & 9.67  & \textbf{21.59} \\
DE-EN   & 20.48 & 19.30 & 10.85 & \textbf{31.81} \\
\rowcolor{alertcolor} 
EN-CS   & 2.37  & 5.80  & 2.87  & \textbf{12.74} \\
\rowcolor{alertcolor} 
EN-DE   & 2.87  & 12.00 & 6.62  & \textbf{24.04} \\
EN-ET   & 2.64  & 6.00  & 2.47  & \textbf{12.59} \\ 
\rowcolor{alertcolor} 
EN-FI   & 2.70  & 3.80  & 3.42  & \textbf{9.41}  \\
\rowcolor{alertcolor} 
EN-RU   & 1.98  & 9.40  & 2.80  & \textbf{12.82} \\
\rowcolor{alertcolor} 
EN-TR   & 2.74  & 6.30  & 3.56  & \textbf{11.70} \\
\rowcolor{alertcolor} 
EN-ZH   & 0.21  & 14.60 & 1.16  & \textbf{22.53} \\
ET-EN   & 15.62 & 14.50 & 2.43  & \textbf{16.85} \\
FI-EN   & 14.16 & 9.70  & 3.79  & \textbf{14.97} \\
RU-EN   & 21.08 & 16.70 & 10.19 & \textbf{22.22} \\
TR-EN   & 14.41 & 14.70 & 4.52  & \textbf{18.47} \\
ZH-EN   & 14.74 & 13.90 & 9.22  & \textbf{15.18} \\
\bottomrule
\end{tabular}%
}
\end{table}

To investigate parameter conflict, we compare \modelname~with three methods using BLEU scores. 
The methods are:
\llama-1B (Base Model), 
SFT (Full Fine-tuned on bilingual corpora in one stage with Base Model),
PT+SFT (Full Fine-tuned in two stages with Base Model: first on monolingual, then on bilingual corpora), 
and \modelname.
Table~\ref{tab:param_conflict_results} presents BLEU scores.
The performance of PT+SFT is particularly revealing.
Despite monolingual exposure in Stage 1, this full-parameter two-stage fine-tuning often fails to benefit the translation task, performing worse than SFT and even the Base Model in several EN-X tasks (e.g., EN-CS, EN-DE, EN-ZH). 
This strongly suggests that without a specialization mechanism, monolingual and bilingual fine-tuning objectives enter into a parameter conflict, where translation gradients may overwrite or interfere with adaptations from monolingual data.
In contrast, \modelname~consistently and significantly outperforms all baselines. 
This superior performance, especially compared to PT+SFT , highlights how its core design effectively mitigates this detrimental parameter interference.

\subsection{The Effect of Expers Number }

\begin{figure}[t]
    \centering
    \includegraphics[width=0.8\linewidth]{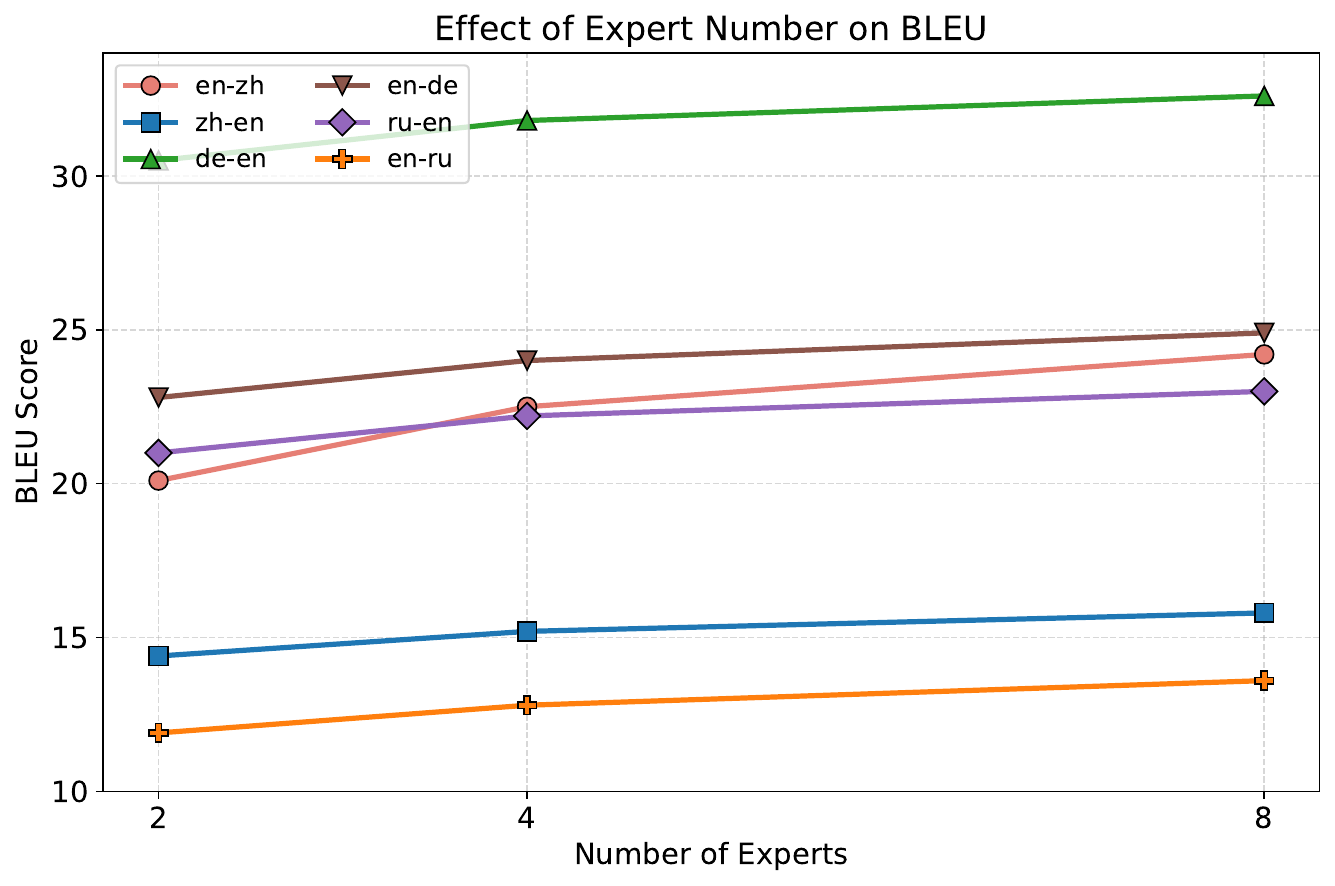} 
    \caption{Effect of Expert Number on BLEU for EN-ZH and ZH-EN Translation Tasks.}
    \label{fig:effect_expert_number}
\end{figure}
We investigated the effects of varying the number of experts within each MoE layer on translation performance. We experiment with 2, 4 (our default setting) and 8 experts for the LM Experts and MT Experts groups.
Figure~\ref{fig:effect_expert_number} shows the BLEU scores obtained in multiple language pairs tasks.

The results show that increasing the number of experts from 2 to 4 and then to 8 leads to better BLEU scores on both translation tasks.
This suggests that a larger number of experts provides a greater capacity for specialization, allowing the model to learn more nuanced representations of the source and target languages.
However, as the number of experts increases, the routing mechanism can become more complex and the computational cost also increases.

We opted for 4 experts per group to balance the potential benefits of increased capacity, computational limitations, and the need for a fair comparison with baseline models.

\subsection{The Effect of \texorpdfstring{$\lambda_{LB}$}{lambda\_LB}}

\begin{table}[t]
\centering
\caption{Effect of $\lambda_{\text{LB}}$ on different metrics. BLEU is the average score of test set. Expert Utilization is the average utilization rate of each expert.}
\renewcommand{\arraystretch}{1.1}
\resizebox{0.8\columnwidth}{!}{
\begin{tabular}{r|cc}
\hline
\bf $\lambda_{\text{LB}}$ &\bf BLEU &\bf Expert Utilization \\
\hline
0 & 15.0 &  0.40   \\
0.001 & 16.5  & 0.28   \\
\rowcolor{lightgray}\bf 0.01 (ours) & \bf 17.5 & \bf 0.26  \\
0.1 & 17.0 & 0.25  \\
1 & 14.0 & 0.25  \\
\hline
\end{tabular}
}

\label{tab:lambda_effect}
\end{table}

The load balancing coefficient $\lambda_{\text{LB}}$ plays a crucial role in the performance and efficiency of our MoE model.
This hyperparameter controls the trade-off between expert specialization and load balancing.
We conducted experiments with different $\lambda_{\text{LB}}$ values to investigate its impact on translation quality and expert utilization.
We use the following $\lambda_{\text{LB}}$ values in our experiments: 0, 0.001, 0.01, 0.1, 1.
All other hyperparameters are kept constant in these experiments.
Table~\ref{tab:lambda_effect} shows the average results and standard deviations for 3 runs for each metric of $\lambda_{\text{LB}}$.

We observe that $\lambda_{\text{LB}}$ increases, expert utilization becomes more balanced, but the BLEU score initially improves and then starts to decline.  
This suggests that a moderate $\lambda_{\text{LB}}$ value is crucial for achieving expert specialization and load balance.  
A value of $\lambda_{\text{LB}}=0.01$  achieves the best balance between translation quality and expert use in our experiments.

\subsection{The Analysis of \texorpdfstring{top-$k$}{top-k}}

In MoEs, the top-$k$, representing the number of experts selected by the gating network to process each input token, plays a pivotal role in balancing model capacity and computational efficiency. 
Numerous studies have shown that increasing $k$ in MoE models generally improves performance \cite{zhu2025overcoming}.
However, increasing $k$ also increases the computational cost of the model.
Selecting a larger $k$ requires routing each token to more experts, resulting in increased FLOPs and memory requirements, particularly during the computationally intensive feedforward and backpropagation phases.

\subsection{Cross-Lingual Generalization and Knowledge Transfer}
 
\begin{figure}[t]
    \centering
    \includegraphics[width=0.7\linewidth]{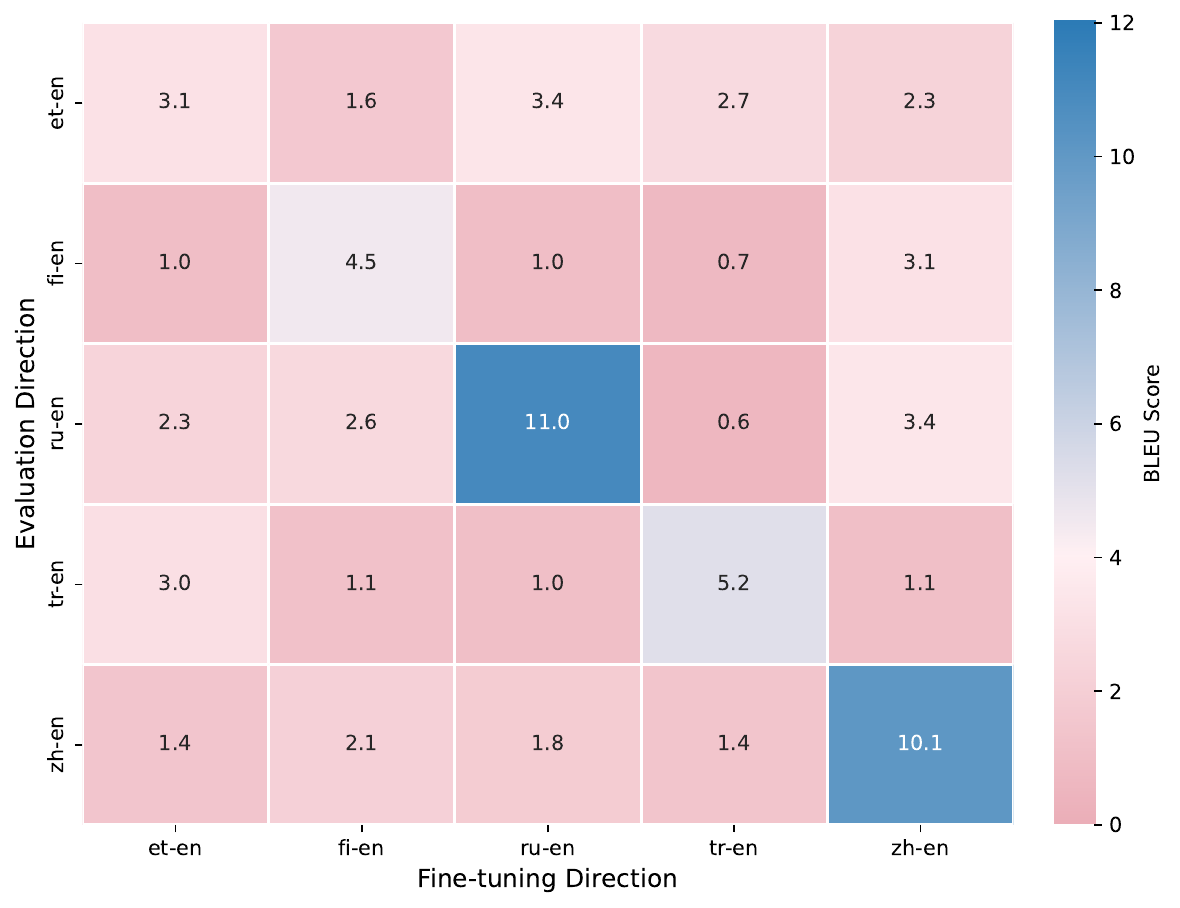}
    \caption{BLEU improvement achieved on other language pairs using the \modelname~after the first stage.}
    \label{fig:scenario-a}
\end{figure}

\begin{figure}[t]
    \centering
    \includegraphics[width=0.7\linewidth]{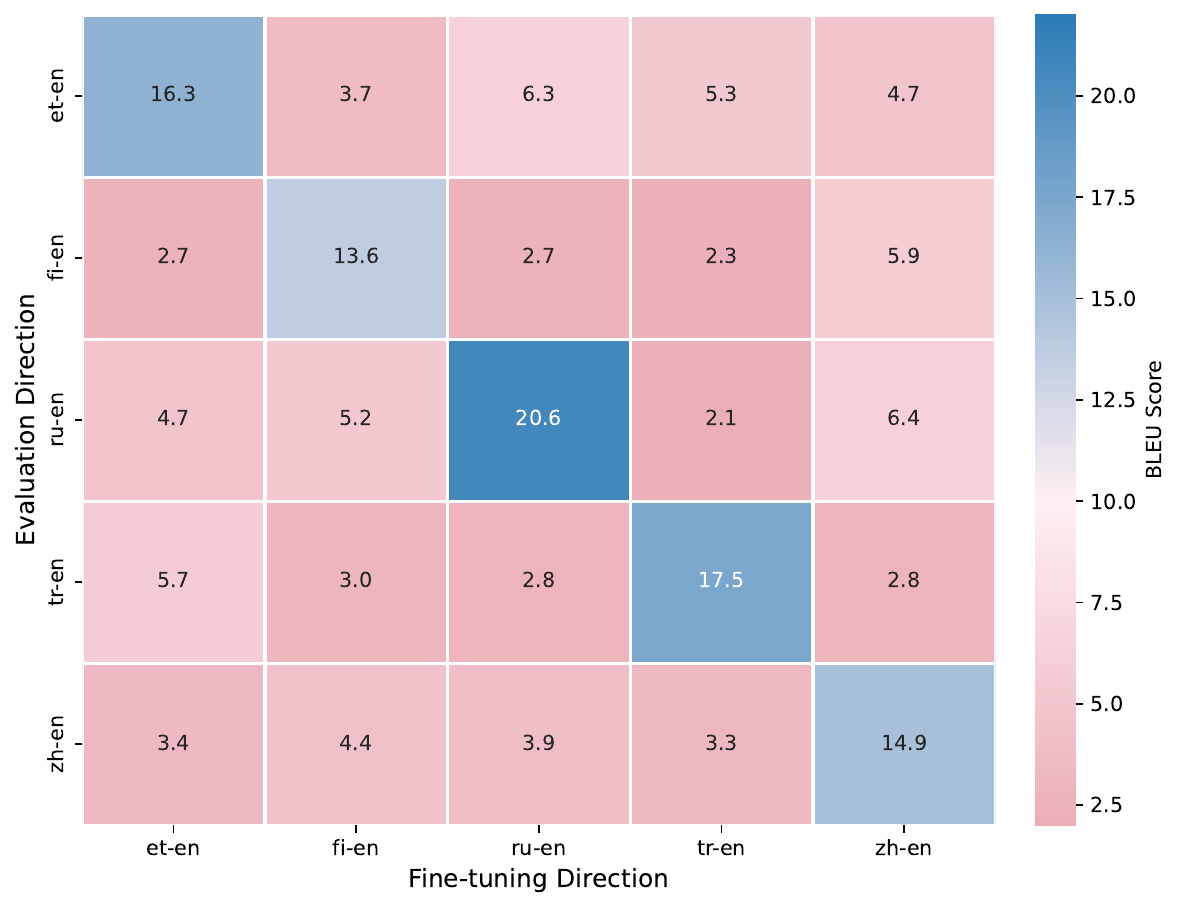}
    \caption{BLEU improvement achieved on other language pairs using the \modelname~after the second stage.}
    \label{fig:scenario-b}
\end{figure}
 
To investigate the cross-lingual generalization capabilities of Mix-MoE and the mechanisms of knowledge transfer between language pairs, we conducted an experiment where the model was fine-tuned on parallel data for a single specific language pair, and subsequently evaluated on a suite of other language pairs without further training. Figure~\ref{fig:scenario-a} illustrates the BLEU score improvements on these unseen target pairs (uneval\_langs) following the first stage of training (Monolingual Expert Training). Figure~\ref{fig:scenario-b} presents the comparative results after the second stage (MT Expert Training), highlighting the incremental gains achieved over Stage 1. The results indicate that fine-tuning Mix-MoE on a single language direction generally leads to performance improvements in other, unseen directions, albeit to varying degrees. This positive transfer suggests that the model is not merely overfitting to the specific fine-tuning pair but is developing more generalized translation capabilities.

\subsection{Scalability Evaluation on Llama3.1-8B and \qwen-14B}

\begin{figure}[t]
\centering
\centerline{\includegraphics[scale=0.7]{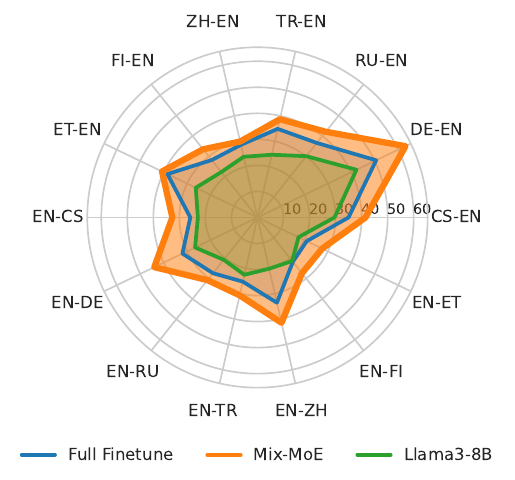}}
\caption{Comparison of BLEU scores across 14 language directions for finetuning the Llama3.1-8B model using full finetuning, and our proposed method.}
\label{fig:llama3_8b}
\end{figure}

\begin{figure}[t]
\centering
\centerline{\includegraphics[scale=0.7]{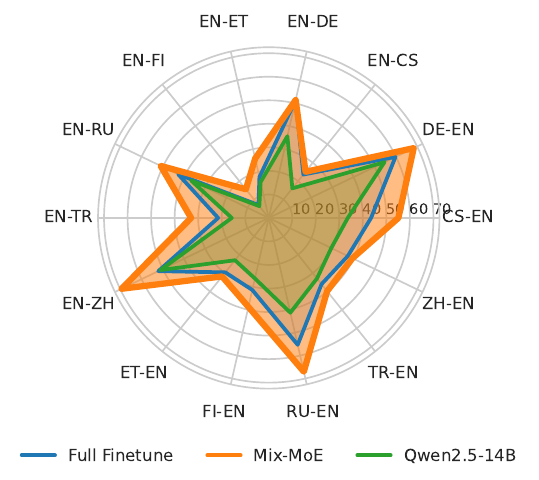}}
\caption{Comparison of BLEU scores across 14 language directions for finetuning the \qwen-14B model using full finetuning, and our proposed method.}
\label{fig:qwen2_5_14b}
\end{figure}

To evaluate the scalability and effectiveness of \modelname~, we extended our experiments to Llama3.1-8B and \qwen-14B.
As illustrated in Figure~\ref{fig:llama3_8b}, the \modelname~method outperforms both the full finetuning approach and the original Llama3.1-8B across all 14 language directions.
This demonstrates the applicability of the \modelname~method for models of different sizes.
Similarly to Llama3.1-8B experiments, we compared \modelname~with full finetuning and the base \qwen~model.
The results, presented in Figure~\ref{fig:qwen2_5_14b}, corroborate the findings of our experiments with LLaMA3.1-8B and other smaller models. The \modelname~method demonstrates significant BLEU score improvements over both full finetuning and the base \qwen-14B model across nearly all language pairs.
These consistent and substantial gains on the \qwen~model further underscore the robustness and scalability of the \modelname~framework.

\section{Conclusion}

In this work, we have presented \modelname, a novel mixed Mixture-of-Experts framework for multilingual MT by finetuning LLMs. 
By strategically dividing MoE layers into monolingual and bilingual experts, and incorporating a routing mechanism enhanced by FFT features from model representations, \modelname~effectively mitigates parameter interference and enhances knowledge transfer. 
Our experimental results demonstrate the superior performance of \modelname~compared to existing baselines, demonstrating its potential to advance the field of multilingual MT with LLM.

\section*{Acknowledgments}

The present research was supported by the National Key Research and Development Program (Grant No.2023YFE0116400), National Natural Science Foundation of China Youth Fund (Grant No.62306210) and the Tianjin Natural Science Foundation of Youth Fund (Grant No.23JCQNJC01690). 
We would like to thank the anonymous reviewers for their insightful comments.

\bibliographystyle{IEEEtran}
\bibliography{custom}

\end{document}